\documentclass[runningheads]{llncs}
\usepackage{graphicx}
\usepackage{comment}
\usepackage{amsmath,amssymb} % define this before the line numbering.
\usepackage{color}

\usepackage{subfigure}
\usepackage{multirow}
\usepackage{multicol}
\usepackage{booktabs}
\usepackage{makecell}
\usepackage{caption}
\usepackage{textcomp}
\newcommand{\tabincell}[2]{\begin{tabular}{@{}#1@{}}#2\end{tabular}}
\usepackage{epstopdf}
\def\etal{{\em et al.~}}

\usepackage{tikz}
\usepackage{epsfig}
\usepackage{threeparttable}
\usepackage{longtable}
\usepackage{rotating}
\usepackage{tabularx}
\usepackage{enumitem}
\usepackage{bbm}
\usepackage{pifont}
\usepackage{arydshln}
\usepackage{indentfirst}  % space

\usepackage{amsfonts}

\usepackage{mathrsfs}
\usepackage{wrapfig}

\newlength\savewidth
\definecolor{Olive_Green}{rgb}{0.0, 0.55, 0.0}
\usepackage[pagebackref=false,breaklinks=true,letterpaper=true,colorlinks,urlcolor = black,  citecolor = Olive_Green, bookmarks=false]{hyperref}

\usepackage{xspace}

\makeatletter
\DeclareRobustCommand\onedot{\futurelet\@let@token\@onedot}
\def\@onedot{\ifx\@let@token.\else.\null\fi\xspace}

\def\@fnsymbol#1{\ensuremath{\ifcase#1\or *\or \dagger\or \ddagger\or
		\mathsection\or \mathparagraph\or \|\or **\or \dagger\dagger
		\or \ddagger\ddagger \else\@ctrerr\fi}}
	
\begin{document}
% \linenumbers
\pagestyle{headings}
\mainmatter
\def\ECCVSubNumber{3492}  % Insert your submission number here

\title{Learning to Predict Salient Faces: A Novel Visual-Audio Saliency Model} % Replace with your title

\begin{comment}
\titlerunning{ECCV-20 submission ID \ECCVSubNumber} 
\authorrunning{ECCV-20 submission ID \ECCVSubNumber} 
\author{Anonymous ECCV submission}
\institute{Paper ID \ECCVSubNumber}
\end{comment}
%******************

%\begin{comment}
\def\@fnsymbol#1{\ensuremath{\ifcase#1\or *\or \dagger\or \ddagger\or
		\mathsection\or \mathparagraph\or \|\or **\or \dagger\dagger
		\or \ddagger\ddagger \else\@ctrerr\fi}}
	
\titlerunning{Learning to Predict Salient Faces: A Novel Visual-Audio Saliency Model}
\author{Yufan Liu\inst{1,2}$^\dagger$\thanks{\scriptsize Yufan Liu, Bing Li, Weiming Hu are with National Laboratory of Pattern Recognition, Institution of Automation, Chinese Academy of Sciences (CASIA), the School of Artificial Intelligence (AI), University of Chinese Academy of Sciences (UCAS) and CAS Center for Excellence in Brain Science and Intelligence Technology (CEBSIT).} 
Minglang Qiao\inst{4}\thanks{\scriptsize Equal contribution.}
Mai Xu\inst{4}\thanks{\scriptsize Corresponding authors: Mai Xu (maixu@buaa.edu.cn), Bing Li (bli@nlpr.ia.ac.cn).} 
Bing Li\inst{1}$^\ddagger$ 
Weiming Hu\inst{1,2,3}
Ali Borji\inst{5}
}

\authorrunning{Yufan Liu et al.}
\institute{$^{1}$National Laboratory of Pattern Recognition, CASIA \\ $^{2}$AI School, University of Chinese Academy of Sciences \quad $^{3}$CEBSIT \\ $^{4}$The School of Electronic and Information Engineering and Hangzhou Innovation Institute, Beihang University \quad $^{5}$MarkableAI Inc.\\
%\email{yufan.liu@ia.ac.cn},\email{\{minglangqiao,maixu\}@buaa.edu.cn},\email{\{bli,wmhu\}@nlpr.ia.ac.cn}
}

%\institute{$^{1}$NLPR, CASIA \& UCAS \& CEBSIT \quad $^{2}$ %Beihang University \\
%	$^{3}$Hangzhou Innovation Institute, Beihang University \quad %$^{4}$MarkableAI Inc.\\

%\end{comment}
%******************
\maketitle

\begin{abstract}
Recently, video streams have occupied a large proportion of Internet traffic, most of which contain human faces. Hence, it is necessary to predict saliency on multiple-face videos, which can provide attention cues for many content based applications.
However, most of  multiple-face saliency prediction works only consider visual information and ignore audio, which is not consistent with the naturalistic scenarios. Several behavioral studies have established that sound influences human attention, especially during the speech turn-taking in multiple-face videos. In this paper, we thoroughly investigate such influences by establishing a large-scale eye-tracking database of Multiple-face Video in Visual-Audio condition (MVVA). Inspired by the findings of our investigation, we propose a novel multi-modal video saliency model consisting of three branches: visual, audio and face. The visual branch takes the RGB frames as the input and encodes them into visual feature maps. The audio and face branches encode the audio signal and multiple cropped faces, respectively. A fusion module is introduced to integrate the information from three modalities, and to generate the final saliency map.
Experimental results show that the proposed method outperforms 11 state-of-the-art saliency prediction works. It performs closer to human multi-modal attention. 
\end{abstract}

\section{Introduction}\label{sec:intro}
\begin{figure}[t]
	\begin{center}
		\includegraphics[width=.96\linewidth]{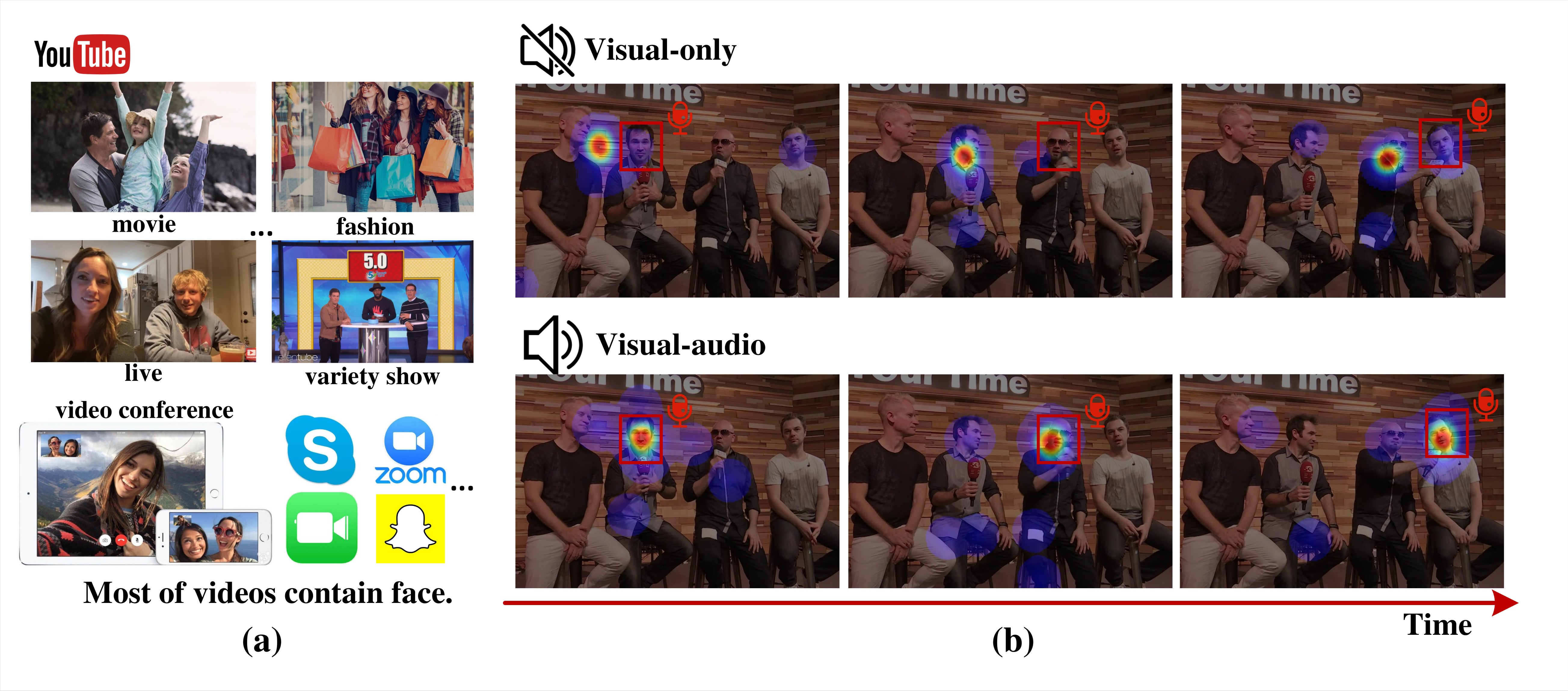}
	\end{center}
	%\vspace{-2em}
	\caption{(a) Thumbnails of various videos over the Internet. Most contain faces.
		(b) Example of visual attention on a multiple-face video. Four persons are speaking in a sequence from the left to the right. The first row (``visual-only'') represents the condition when subjects view only mute frames. The second row (``visual-audio'') shows the condition when both visual and audio information is present.}
	\label{fig:intro}
	%\vspace{-2em}
\end{figure}

Saliency prediction \cite{borji2018saliency} is an effective way to model the deployment of possible attention on visual inputs in biological vision system. %It has been extensively studied within various disciplines, including psychology, cognitive vision, computer vision, and robotics.
In the recent years, a surge of interest in video saliency prediction has emerged, partly because of a large number of its applications in various areas. Besides, it can be also found that most videos over the Internet contain faces, as shown in Fig. \ref{fig:intro}(a). In particular, video conference applications (e.g., Skype and Zoom) have become popular recently, in which almost every frame has human faces. It has been reported \cite{zoomdata} that Zoom Video Communications achieved over 39 billion annualized meeting minutes in 2018. 
Thus, it is necessary to predict saliency on multiple-face videos, since saliency can be used as attention cues for the content based applications, including perceptual video coding \cite{xu2018find}, quality assessment \cite{li2018bridge} and panoramic video processing \cite{xu2018predicting}.

Most of the video saliency works focus on the visual information and few works have taken auditory information into account. Previous works barely mention soundtracks or explicitly discard them during the eye-tracking experiments.
In practice, videos are always played with sound and the world we live in always contains multi-modal information. Human attention is driven by several factors. Two most important ones include visual and auditory cues. As shown in Fig. \ref{fig:intro}(b), humans focus on different regions in visual-only condition vs. visual-audio condition. They fixate at the salient face and transit to another faster when sound is available. Without sound, people can only rely on visual cue (e.g. motion) to locate the speaking person, leading to slower attention transition.
Thus, only considering visual information is not enough to predict where people look.

To address the above shortcomings, we first create a large-scale eye-tracking database dubbed Multiple-face Videos in Visual-Audio condition (MVVA). It includes fixations of 34 subjects viewing 300 videos with diverse content. To the best of our knowledge, so far this is the largest dataset of its kind. During the eye-tracking experiment, both video and audio have been presented to the viewers. Through analysis on our database, we find that faces indeed explain the majority of fixations. We further confirm that audio influences the fixation distribution on faces and attention transition across faces. In particular, human attention in visual-audio condition significantly differs from visual-only condition, when turn-taking takes place.
Inspired by these findings, we propose a novel multi-modal network to predict fixations on videos in the visual-audio condition. Our work takes faces, global visual content and audio information into consideration. It consists of three branches, namely visual, audio and face branches, to process these information respectively. \textcolor{black}{Specifically, the visual branch constructs a two-stream architecture to model spatial-temporal visual saliency representation. Without other information, the output of the visual branch can be seen as the saliency map under the visual-only condition.} The audio branch encodes the 1D audio signal into a 2D feature map sequence. 
Additionally, the face branch processes multiple cropped faces and explore the relationship between them. And then it generates a face saliency map. After that, a fusion module is introduced to integrate the three modalities, and to generate the final saliency map. We study the impact of each of these cues individually. 

\begin{table}[t]
	%\vspace{-2.8em}
	\centering
	\begin{small}
		\renewcommand{\tabcolsep}{3pt}
		\caption{The information of the existing visual-audio eye-tracking databases.}
		%\vspace{-.1em}
		\resizebox{0.98 \textwidth}{!}{
			\begin{tabular}{|c|ccccc|}
				\hline
				\textbf{Database} & \textbf{Video Num.} & \textbf{Resolution} & \textbf{Duration} & \textbf{Subject} & \textbf{Details} \\
				\hline
				\tabincell{c}{Coutrot I \\ \cite{coutrot2013toward}} & 60 & $720\times576$ & $10$-$24.8$sec & \tabincell{c}{20 (10 per\\ auditory condition)} & \tabincell{c}{French; Scenes: one MO (Moving Object),\\ several MO, conversation and landscapes.}\\
				\hline
				\tabincell{c}{Coutrot II \\ \cite{coutrot2014saliency}} & 15 & $720\times576$ & $12$-$30$sec & \tabincell{c}{72 (18 per\\ auditory condition)} & French; Scenes: Conversation.\\
				\hline
				\tabincell{c}{Coutrot III \\ \cite{coutrot2015efficient}} & 15 & $1232\times504$ & $20$-$80$s & 40 & English; Scenes: 4 persons meeting. \\
				\hline
				\tabincell{c}{Pierre \etal \\ \cite{marighetto2017audio}} & \tabincell{c}{148 (from Coutrot\\ dbs \& Hollywood)} & $ \leq1232\times504$ & $0.9$-$35$s & \tabincell{c}{averaged 44 \\ each experiment}  & Scenes: MO, conversation and landscapes. \\
				\hline
				\textbf{Ours} & 300 & $\geq1280\times720$ & $10$-$30$s & 34 & Chinese \& English; 6 kinds of scenes \\
				\hline
			\end{tabular}
		}
		\label{tab:dataset_summary}
		%\vspace{-1.9em}
	\end{small}
\end{table}

To summarize, our main contributions include:
\begin{itemize}
	%\vspace{-.8em}
	\item We introduce a large-scale eye-tracking database including multiple-face videos with sound, to facilitate the research on visual-audio saliency prediction.
	\item We present a thorough analysis on our database and study how human attention is affected by multiple factors including face and sound.
	\item We propose a novel multi-modal network, which fuses visual, face and audio information to obtain effective features for accurate saliency prediction.
	%\vspace{-1em}
\end{itemize}

%\vspace{-.6em}
\section{Related Work} \label{sec:related_work}
%\vspace{-.5em}

\noindent\textbf{Visual saliency prediction}.
Visual saliency models have been widely developed to predict where people look in images ~\cite{huang2015salicon,zhang2016exploiting,pan2017salgan,wang2017deep,cornia2018predicting} or videos \cite{hossein2015many,bak2017spatio,liu2017predicting,jiang_OMCNN,wang2018revisiting,min2019tased,zanca2019gravitational}. Recently, DNNs have achieved a great success in visual saliency prediction. 
Over images, some deep saliency models \cite{huang2015salicon,wang2017deep} use multi-scale visual information to predict saliency. 
Over videos, most works \cite{jiang_OMCNN,wang2018revisiting,liu2017predicting} combine a CNN and an LSTM to learn spatial and temporal visual features. Bak \etal \cite{bak2017spatio} proposed a two-stream CNN architecture. RGB frames and optical flow sequences were fed to the two streams.
Zanca \etal \cite{zanca2019gravitational} leveraged various visual features, such as face and motion, to predict the fixation scanpath.
Recently, some works have focused on predicting saliency over multiple-face videos. Liu \etal \cite{liu2017predicting} presented an architecture which combined a CNN and a multiple-stream LSTM to learn face features. 
None of the methods above take audio modality into account. In contrast, our approach utilizes both audio and video modalities.

\noindent\textbf{Visual-audio saliency prediction}. 
Only a few methods take the auditory information into account.
Early saliency models adopted hand-crafted features. For instance, in \cite{coutrot2014audiovisual}, low-level features (e.g., luminance information) and faces are used as visual information. Audio is fed into a speaker diarization algorithm  to locate the speaking person. A saliency map is then generated by integrating the two modalities. \cite{coutrot2015efficient} improved this method by taking the body into consideration. These methods rely heavily on the detection algorithms, which limits their performance and usability.
Recent works tend to make use of learning-based methods.
Tsiami \etal \cite{tsiami2016towards} combined a visual saliency model \cite{itti1998model} and an audio saliency model \cite{kayser2005mechanisms}. But it only considers the scenario that a simple stimuli moving in clustered images. 
More recently, \cite{tavakoli2019dave} used a two-stream 3D-CNN to encode visual and audio information into feature vectors, which are then concatenated to learn the final prediction.

\noindent\textbf{Visual-audio databases}.
Few datasets have been collected for studying visual-audio attention as shown in Tab. \ref{tab:dataset_summary}. They have three main drawbacks. Firstly, they usually have a small scale. The number of videos in these datasets are typically under 100. Secondly, they contain only one or a few scenes. For example, Coutrot II \cite{coutrot2014saliency} and Coutrot III \cite{coutrot2015efficient} only consider eye-tracking events in a specific scene. Thirdly, their videos have low resolution. Coutrot I \cite{coutrot2013toward} and Coutrot II \cite{coutrot2014saliency} contain videos with a 720 x 576 resolution. Consequently, the existing visual-audio saliency prediction methods are designed under specific conditions (e.g. under a certain scene or a low resolution). The efficiency and generalization of these models need further verification. Driven by these motivations, here we propose a dataset of 300 videos with the resolution of at least 1280 x 720 over 6 different scenes. Further, we analyze our dataset to reveal the impact of audio on human attention, and give some inspirations for saliency prediction.

%\vspace{-4pt}
\section{The Proposed Dataset}
%\vspace{-5pt}
In this section, we introduce a large-scale eye-tracking database called Multiple-face Video in Visual-Audio condition (MVVA). The proposed dataset contains eye-tracking fixations when both audio and video were presented. To the best of our knowledge, our dataset is the first public eye-tracking database that has multiple-face videos with audio. In addition to saliency, it can be used in other research areas such as sound localization, since the faces of speakers are manually marked in our dataset. 
Our dataset is publicly available in \emph{https://github.com/MinglangQiao/MVVA-Database}.

%\vspace{-3pt}
\subsection{Data collection}
%\vspace{-6pt}

\noindent\textbf{Stimuli.} A total number of 300 videos with 146,529 frames, containing both images and audio, were collected. Among them, 143 videos were selected from MUFVET \cite{liu2017predicting} and other 157 videos were selected from YouTube, with the criterion that the videos should contain obvious faces and audio. All of them were encoded by H.264 with duration varying from 10 to 30 seconds. Note that these videos are either indoor or outdoor scenes, and can be classified into 6 categories: TV play/movie, interview, video conference, variety show, music and group discussion.
The audio content covers different scenarios including quiet scenes (e.g., news broadcasting) and noisy scenes (e.g., interview at subway).

\noindent\textbf{Apparatus.} For monitoring the binocular eye movements, an eye tracker, EyeLink 1000 Plus \cite{eyeLink1000}, was used in our experiment. EyeLink1000 Plus is an integrated eye tracker with a 23.8'' TFT monitor at screen resolution of 1280x720. During the experiment, EyeLink1000 Plus captured gaze data at 500 Hz. According to \cite{eyeLink1000}, the gaze accuracy can reach 0.25-0.5 visual degrees in the head free-to-move mode. For more details on EyeLink1000 Plus, see \cite{eyeLink1000}.

\begin{figure}[t]
	\centering
	%\vspace{-1.2em}
	\hspace{-1.2em}
	\subfigure[]{
		\hspace{-3.2em}
		\begin{minipage}[t]{0.6\linewidth}
			\centering
			\label{fig:faceFeature_nss}
			\includegraphics[width=1.8in]{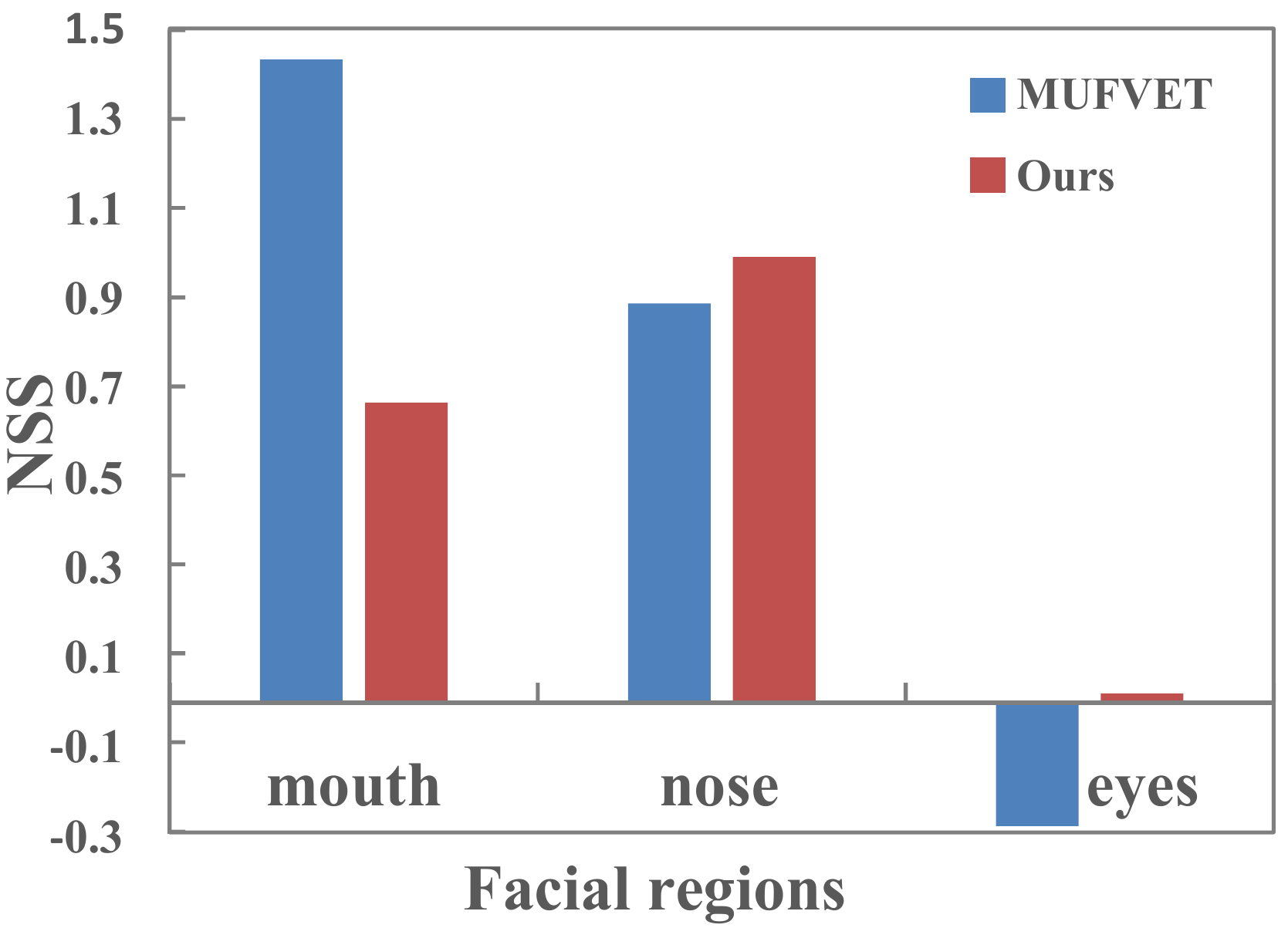}
			%\caption{fig1}
		\end{minipage}%
	}%
	\subfigure[]{
		\hspace{-4.2em}
		\begin{minipage}[t]{0.6\linewidth}
			\centering
			\label{fig:contextual_NSS}
			\includegraphics[width=1.8in]{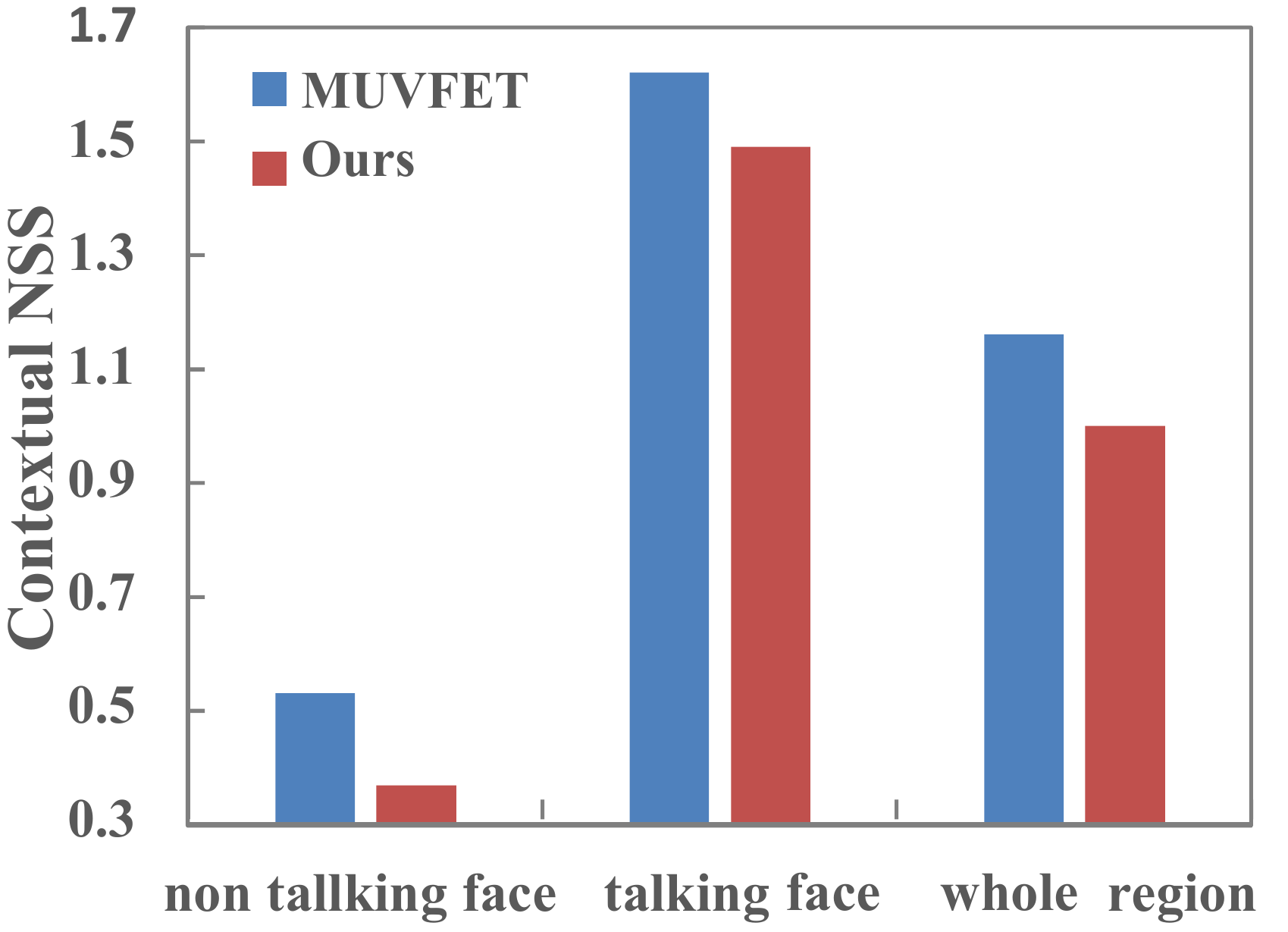}
			%\caption{fig2}
		\end{minipage}%
	}%
	\centering
	%\vspace{-1.4em}
	\caption{(a) NSS of saliency on different facial landmarks in visual-only (MUVFET)/visual-audio (Ours) conditions.
		(b) Contextual NSS of optical flow maps over different face regions.}
	%\vspace{-1.8em}
\end{figure}

\noindent\textbf{Participants.} 34 participants (21 males and 13 females), aging from 20 to 54 (24 in average), were recruited to participate in the eye-tracking experiment.
All participants had normal or corrected-to-normal vision. It is worth pointing out that only subjects who passed the eye tracking calibration were quantified for the experiment. As a result, 34 subjects (out of 39) were selected in our experiment.

\noindent\textbf{Procedure.} During the eye tracking experiment, all subjects were required to sit on a comfortable chair with the viewing distance of $\sim 55cm$ from the screen. Before viewing the videos, each subject was required to perform a 9-point calibration for the eye tracker. Afterwards, videos were shown in a random order and subjects were asked to view them freely. Note that the audio and video stimuli were presented simultaneously during the experiment. In order to avoid eye fatigue, the 300 videos were equally divided into 6 sessions, and there was a 5-minute rest after viewing each session. Besides, a 5-second blank period with a black screen was inserted between each two successive videos for a short break. In total we collected 5,013,980 fixations over all 34 subjects and the 300 videos.

%\vspace{-3pt}
\subsection{Database analysis}
%\vspace{-3pt}
Here, we thoroughly analyze our data. To annotate faces and face landmarks in video frame, we used \cite{zhang2016joint} and \cite{ren2014face}, respectively, and then corrected the predictions manually. The talking/non-talking faces are manually annotated.

\begin{figure}[t]
	\begin{minipage}{0.6\linewidth}
	\centering
		\vspace{-.6em}
		\includegraphics[width=1.01\linewidth]{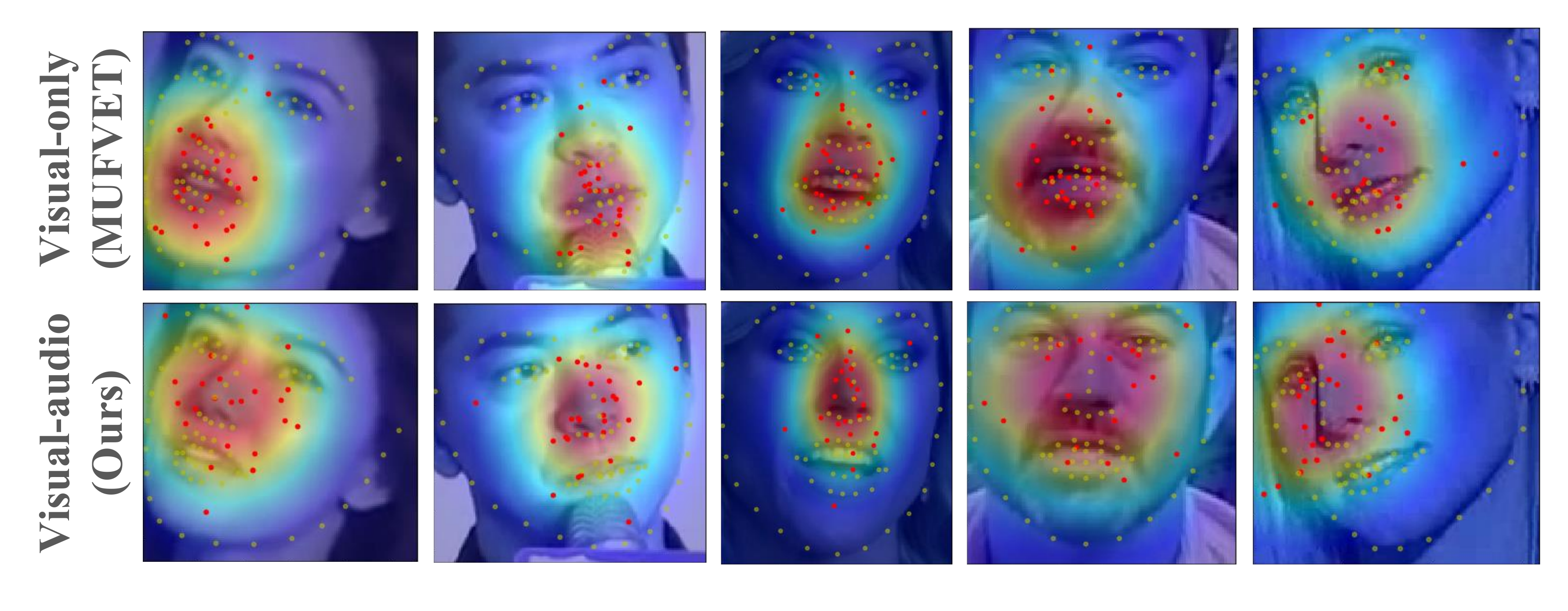}
	\end{minipage}
    \hfill
	%\vspace{-2.5em}
	\begin{minipage}{0.4\linewidth}
	\caption{Examples of saliency maps in visual-only (the first row) and visual-audio condition (the second row). The red dots are fixation points, and the yellow dots are facial landmarks.}\label{fig:attention_example}
    \end{minipage}
	%\label{fig:attention_example}
	%\vspace{-2em}
\end{figure}

\noindent\textbf{Finding 1: Audio influences the fixation distribution on faces.}
With the presence of audio, fixation distribution is different from that of visual-only scenario.
First, we find that the face saliency distribution in visual-audio condition is slightly more dispersed than that in visual-only condition. We compute the averaged entropy and dispersion \cite{marighetto2017audio,coutrot2012influence} of each face saliency map, and obtain 10.58 and 44.06 on our MVVA (visual-audio condition), larger than 10.16 and 39.34 of MUVFET (visual-only condition). It may be because people need to focus on mouth to identify the talking face without audio, but do not need that when audio is available. Second, as shown in Fig.~\ref{fig:attention_example}, in the visual-audio condition, human attention tends to fixate at the center of the face (i.e., near the nose), while people tend to focus on mouth in the visual-only condition. We calculate the Normalized Scanpath Saliency (NSS) between saliency map and different facial landmarks to quantify the correlation between salient regions and facial regions in Fig. \ref{fig:faceFeature_nss}. It depicts that saliency maps in our database have the highest NSS values on nose, while on  MUFVET the salient region is on mouth.
This may be because people do not need to concentrate on the mouth motion, when they can clearly hear the sound. Third, attention transits from mouth/nose to eyes when face becomes larger.
We compute NSS of saliency map on facial landmarks, and calculate the Pearson correlation coefficient between the NSS and the normalized face size. We find that the Pearson correlation coefficients between face size and NSS on \{eyes, mouth, nose\} in order are \{0.29, -0.44, -0.12\} in our dataset, and \{0.54, -0.49, 0.14\} in MUFVET. Positive correlation between face size and NSS on eyes reflects more attention on eyes when subjects are viewing larger faces.

\begin{figure}[t]
	\centering
	%\vspace{-1.2em}
	\hspace{-1.2em}
	\subfigure[]{
		\hspace{-3.6em}
		\begin{minipage}[t]{0.61\linewidth}
			\centering
			\label{fig:finding3_vis}
			\includegraphics[width=2.2in]{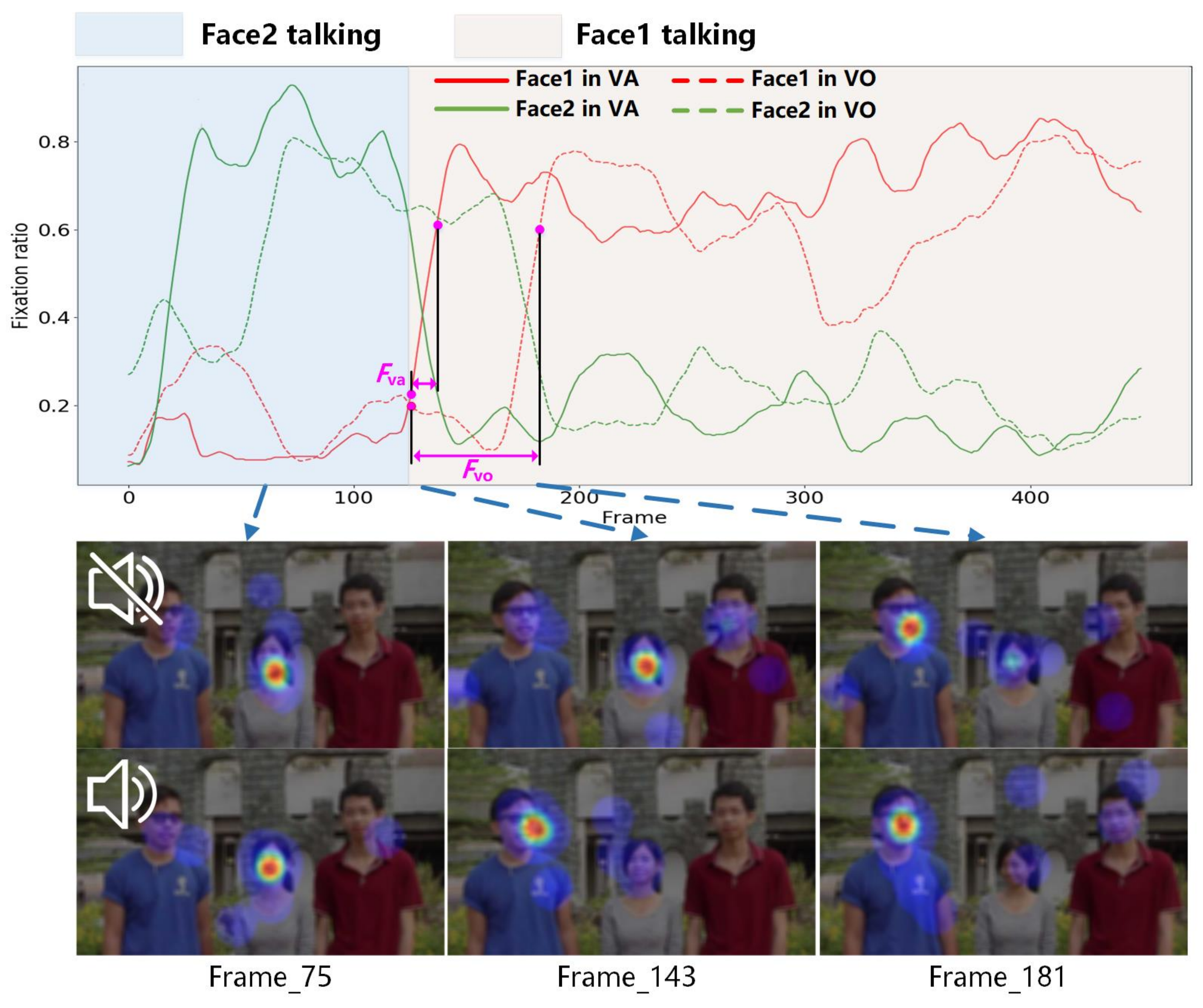}
			%\caption{fig1}
			\vspace{.25em}
		\end{minipage}%
	}%
	\subfigure[]{
		\hspace{-4.6em}
		\begin{minipage}[t]{0.6\linewidth}
			\centering
			\label{fig:face_opticalFlow}
			\includegraphics[width=2.65in]{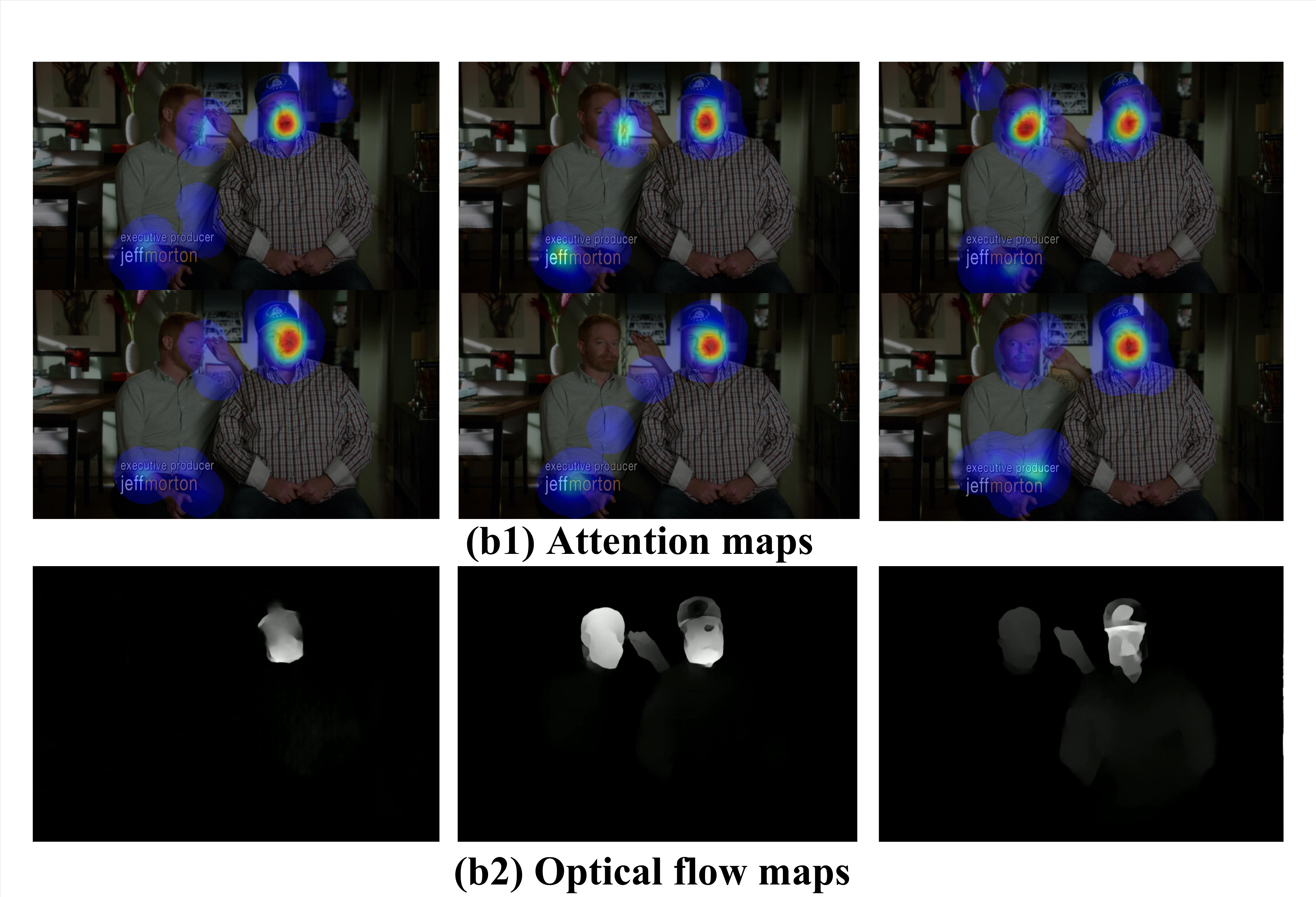}
			%\caption{fig2}
		\end{minipage}%
	}%
	\centering
	%\vspace{-1.6em}
	\caption{(a) An example of attention transition in Visual-Only (VO, the first row of heat maps) and Visual-Audio condition (VA, the second row of heat maps).
		(b1) One video example showing the saliency difference between visual-only condition (the first row) and visual-audio condition (the second row). The person at the right is talking while the other is turning his head. (b2) The corresponding optical flow maps of each frame.}
	%\vspace{-2em}
\end{figure}

\noindent\textbf{Finding 2: In the turn-taking scenes, the transition of fixations across faces is largely influenced by audio.} 
Fig. \ref{fig:finding3_vis} shows an example of attention transition in the turn-taking scenes. It can be observed that human fixations transit and follow the talking face faster in the visual-audio condition than that in the visual-only condition.  Fig. \ref{fig:intro} also shows the similar observation. 
For quantitative analysis, we compare the attention transition time in visual-audio and visual-only conditions. We define the attention transition time by the average number of frames that fixations transit to the talking face, when turn-taking happens. 
Here, $F_{\text{va}}$ and $F_{\text{vo}}$ denote the attention transition time in MVVA (visual-audio condition) and MUVFET (visual-only condition), respectively. The results of $F_{\text{va}}$ and $F_{\text{vo}}$ are 30 and 24 frames. Thus, the attention transition time in visual-audio condition is shorter than that in visual-only condition by $25\%$.
From the above results, we can conclude that the fixations transit across faces are largely influenced by audio.

\noindent\textbf{Finding 3: Human attention is more influenced by motion in the absence of audio.} 
It is intuitive that people are guided by the visual cues (e.g., motion) more in the visual-only condition, compared to the visual-audio condition. This is because people can only rely on the visual cues to figure out what is going on in the video under the visual-only condition.
For instance, in Fig. \ref{fig:face_opticalFlow}, in visual-only condition attention is mostly attracted to the person on the left who is turning his head, while in the visual-audio condition, subjects concentrate on the right speaking person.
To quantify the relationship between motion and saliency, we computed the contextual NSS \cite{tavakoli2019dave} of the optical flow maps on fixations. Fig. \ref{fig:contextual_NSS} illustrates that human attention correlates more with motion in the visual-only condition.

%\vspace{-6pt}
\section{The Proposed Method}
\begin{figure*}[t]
	\begin{center}
		%\vspace{-.6em}
		\includegraphics[width=.96\linewidth]{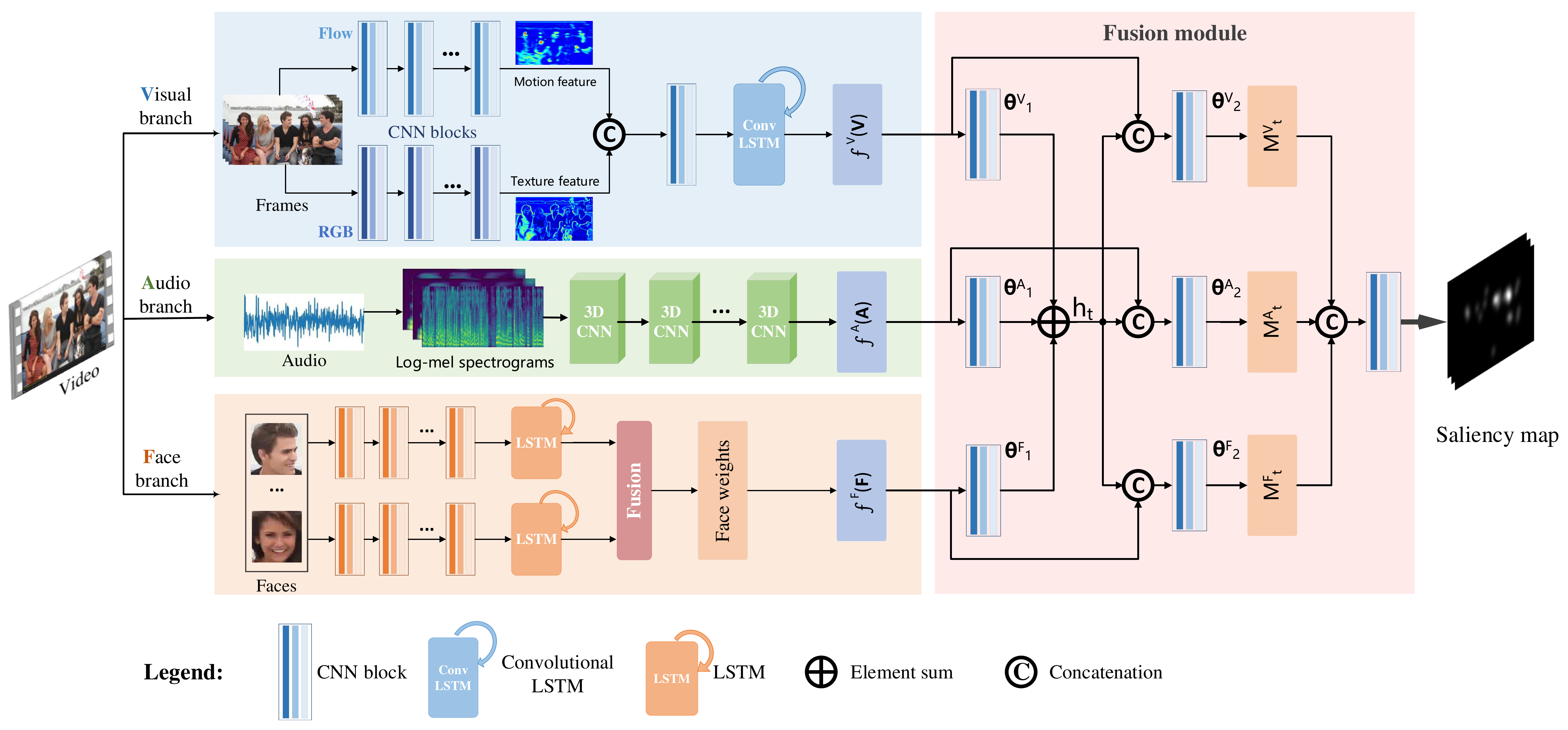}
	\end{center}
	%\vspace{-2em}
	\caption{Overall framework of the proposed method.}
	\label{fig:framework}
	%\vspace{-1.6em}
\end{figure*}

%\vspace{-6pt}
According to the findings above, visual information, audio and faces are all important factors that influence human attention.
In this section, we introduce our multi-modal saliency method that utilizes these information for predicting fixations over multiple-face videos. 
Fig. \ref{fig:framework} summarizes the overall framework of the proposed method. A three-branch neural network is used to integrate multiple information cues and to generate a saliency map.
Particularly, a video segment $Video = \{\mathbf{V}, \mathbf{A}, \mathbf{F}\}$, comprising visual frames $\mathbf{V}=\{V_t\}_{t=1}^T$, audio signals $\mathbf{A}=\{A_t\}_{t=1}^T$ and faces $\mathbf{F}=\{F_t\}_{t=1}^T$, is first fed into our multi-modal neural network. Each component of the video segment is conveyed to the corresponding branch of the network. The predicted saliency maps $\mathbf{S}=\{S_t\}_{t=1}^T$ are then computed as:
\begin{flalign}\label{equ:overall}
\begin{aligned}
\mathbf{S} = f(\mathbf{V, A, F}) = \Phi(f^V(\mathbf{V}), f^A(\mathbf{A}), f^F(\mathbf{F})),
\end{aligned}
\end{flalign}
where $f(\cdot)$ is the proposed model, and $f^V(\cdot), f^A(\cdot), f^F(\cdot)$ are the three branches for visual, audio and face cues, respectively. Besides, $\Phi(\cdot)$ is the fusion module to integrate the three modalities and to generate the final saliency maps.

%\vspace{-6pt}
\subsection{Architecture}

\noindent\textbf{Visual branch}. \textcolor{black}{Fig. \ref{fig:framework} shows visual branch constructs a two-stream CNN \& convolutional LSTM architecture to model spatial-temporal visual representation. 
In detail, on the one hand, the frames $\mathbf{V}$ are fed to an RGB sub-branch to obtain the features of texture. On the other hand, frames are fed to a flow sub-branch to get the features of motion. Note that the flow sub-branch is initialized by FlowNet~\cite{dosovitskiy2015flownet} so that it can obtain motion-oriented features.}
Then, these extracted features are concatenated (denoted as $\mathrm{C}(\cdot)$) and are fed to a two-layer convolutional LSTM \cite{xingjian2015convolutional}, which is leveraged to process spatial-temporal information. After that, feature maps $f^V(\mathbf{V})$ are obtained as:
\begin{flalign}\label{equ:visual}
\begin{aligned}
f^V(\mathbf{V}) = \mathrm{LSTM}(\mathrm{C}(g_1(\mathbf{V}), g_2(\mathbf{V}))).
\end{aligned}
\end{flalign}
Note that $g_1(\cdot)$ represents the RGB sub-branch, consisting of four CNN blocks of VGG-16 \cite{simonyan2014very}. And $g_2(\cdot)$ denotes the flow sub-branch, which comprises three CNN blocks and one deconvolutional layer of FlowNet.

\noindent\textbf{Audio branch}.
\textcolor{black}{In audio branch, a frequency domain based 3D-CNN is designed to convolute 1D audio signal by converting it to 2D spectrum. As such, the spectrum can be better integrated with 2D image features.} 
In detail, the audio signal is first re-sampled to 22kHz and is then transformed to log-mel spectrogram using Short-Time Fourier Transform (STFT) and mel-mapping~\cite{hershey2017cnn}, with a hop length of 512. To be consistent with the visual frame, the log-mel spectrogram is converted into a sequence of successive overlapping frames, and is cropped in a $(-230, 230]$ ms window.
After that, 4-layer 3D-CNNs $g_{3d}(\cdot)$ are embedded to encode the log-mel spectrogram sequence and to obtain the audio feature maps:
\begin{flalign}\label{equ:visual}
\begin{aligned}
f^A(\mathbf{A}) = g_{3d}(\mathrm{STFT}(\mathbf{A})).
\end{aligned}
\end{flalign}

\begin{figure}[t]
	\centering
	%\vspace{-.6em}
	\hspace{-.6em}
	\subfigure[]{
		\hspace{-3.2em}
		\begin{minipage}[t]{0.6\linewidth}
			\centering
			\label{fig:face_branch}
			\includegraphics[width=2.4in]{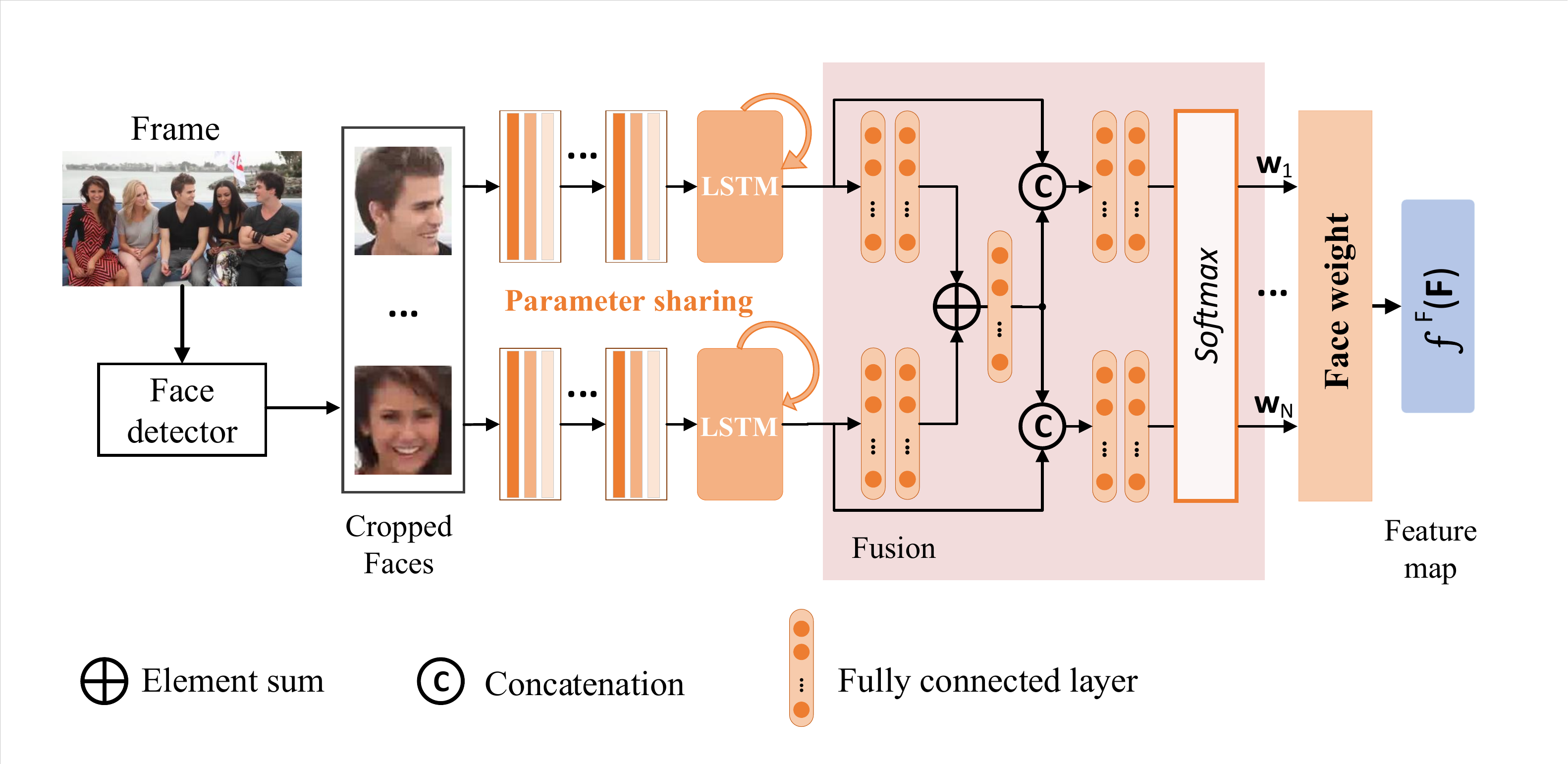}
			%\caption{fig1}
		\end{minipage}%
	}%
	\subfigure[]{
		\hspace{-3.8em}
		\begin{minipage}[t]{0.6\linewidth}
			\centering
			\label{fig:faceNum_example}
			\includegraphics[width=2.35in]{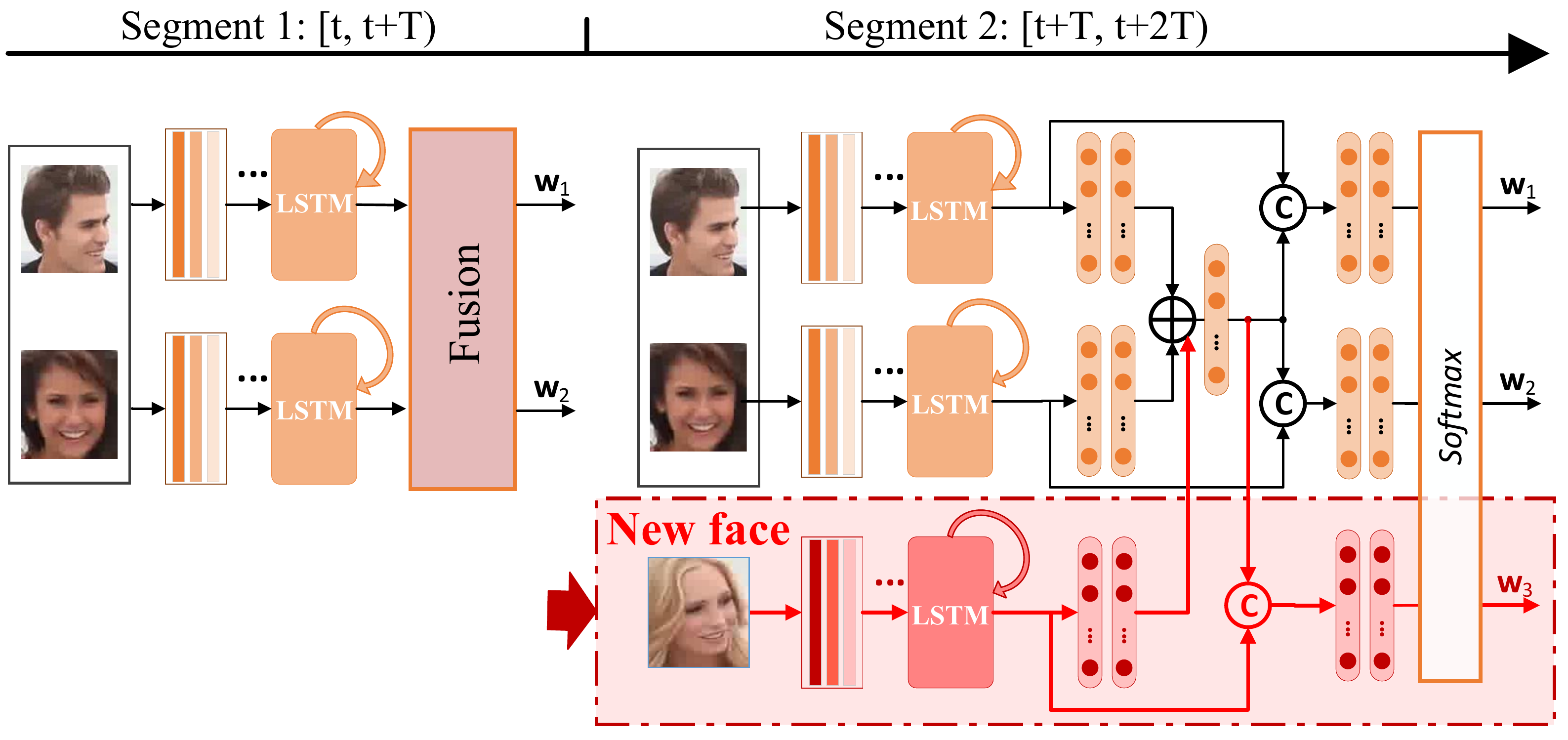}
			\vspace{.3em}
			%\caption{fig2}
		\end{minipage}%
	}%
	\centering
	%\vspace{-1.6em}
	\caption{(a) Structure of face branch.
		(b) An example of face branch processing variant face numbers.}
	%\vspace{-1.6em}
\end{figure}

\noindent\textbf{Face branch}.
\textcolor{black}{In face branch, a dynamic multi-stream spatial-temporal LSTM model is designed for exploring relationship between multi-faces with features interacting with each other.} 
Fig. \ref{fig:face_branch} gives a detailed illustration of the face branch. Firstly, given a sequence of video frames, the MTCNN face detector \cite{zhang2016joint} is leveraged to detect and crop faces. Secondly, $N$ cropped faces are fed into $N$ parameter-shared sub-branches containing an 13-layer CNNs and a 2-layer LSTM, and are transformed to $N$ feature vectors. After that, these features are fused by the fusion part of face branch, which helps face features to capture the correlation and competition with each other. Hence, each face sub-branch perceives the sufficient information and we can obtain $N$ face saliency weights: $\mathbf{w}_1 = \{w_{1,t}\}_{t=1}^T, \mathbf{w}_2, ..., \mathbf{w}_N$. Larger weight for a face means that it is more salient.
Finally, we calculate the face feature map $f^F(F_t)$ at the $t$-th frame as follows,
{\setlength\abovedisplayskip{2pt}
	\setlength\belowdisplayskip{2pt}
	\begin{flalign}\label{equ:GMM}
	\begin{aligned}
	f^F(F_t) = \sum_{n=1}^N w_{n,t} \cdot \mathcal{N}_{n,t}.
	\end{aligned}
	\end{flalign}}
Here, we follow \cite{liu2017predicting} to regard saliency on the $n$-th face as a Gaussian distribution $\mathcal{N}_{n,t}(\mu_{n,t}, \Sigma_{n,t})$\footnote{$\mathcal{N}_{n,t}(\mathbf{x})=\mathrm{exp}\{-\frac{1}{2}(\mathbf{x}-\mu_{n,t})^T\Sigma_{n,t}^{-1}(\mathbf{x}-\mu_{n,t})\}$}. 

The parameter-sharing architecture can process videos with different face numbers. As shown in Fig. \ref{fig:faceNum_example}, a new CNN-LSTM stream is instantiated when there is a new face appearing in the video. To be specific, we use PyTorch to instantiate CNN-LSTM streams with different number at each iteration.

In the training process, firstly we pre-train the face branch. The fixation proportion of the $n$-th face to all faces at frame $t$ (denoted as $w_{n,t}$) is taken as the Ground Truth (GT) weight to supervise the predicted face saliency weight (denoted as $\hat{w}_{n,t}$). Hence, the optimization can be formulated as
{\setlength\abovedisplayskip{2pt}
	\setlength\belowdisplayskip{-2pt}
	\begin{flalign}\label{equ:face_weight_opt}
	%\vspace{-.6em}
	\begin{aligned}
	& \mathrm{min} \sum_{t=1}^T\sum_{n=1}^N ||\hat{w}_{n,t}-w_{n,t}||_2^2,
	\quad s.t. \sum_{n=1}^N \hat{w}_{n,t} = 1.\\
	\end{aligned}
	\end{flalign}}

\noindent\textbf{Fusion}. After encoding each video modality to feature maps, the proposed model integrates visual, audio and face feature maps together to learn a joint representation. 
We propose a fusion module depicted in Fig. \ref{fig:framework}, instead of direct concatenation. 
Given visual, audio and face feature maps $\{f^F(V_t), f^F(A_t), f^F(F_t)\}$, the fusion module performs the computations below:
\begin{flalign}\label{equ:visual_map}
\begin{aligned}
\mathrm{M}^V_t &= \Theta_2^V * \mathrm{C}(h_t, f^V(V_t)), \\
\mathrm{M}^A_t &= \Theta_2^A * \mathrm{C}(h_t, f^A(A_t)), \\
\mathrm{M}^F_t &= \Theta_2^F * \mathrm{C}(h_t, f^F(F_t)), \\
s.t. \quad h_t =\Theta_1^V&*f^F(V_t) + \Theta_1^A*f^F(A_t) + \Theta_1^F*f^F(F_t).  \\
\end{aligned}
\end{flalign}
Note that the $\Theta$s are the parameters of different CNN blocks, \textcolor{black}{which align multi-modal features with different scales and receptive fields (e.g., visual branch outputs global features while face branch outputs local features).}
And '$*$' denotes convolution operator and $\mathrm{C}(\cdot)$ is the concatenation operation. With help of the fusion module, the three branches can share information and preserve original characteristics of themselves.

%\vspace{-5pt}
\subsection{Optimization}
%\vspace{-5pt}
To train and optimize the proposed multi-modal network, we use the GT fixation map $\mathbf{G}$, obtained from the fixation density map, to supervise the predicted saliency map $\mathbf{S}$. The loss function is the Kullback-Leibler (KL) divergence between the two maps,
\begin{flalign}\label{equ:loss_s}
\begin{aligned}
\mathbf{L} = \sum_{t=1}^T KL(G_t||S_t)=\sum_{t=1}^T \sum_{i\in \mathbf{I}} G_t(i) \mathrm{log}\frac{G_t(i)}{S_t(i)},
\end{aligned}
\end{flalign}
in which $i$ denotes a position in the 2D saliency map. Note that KL divergence is chosen because Huang \etal \cite{huang2015salicon} have proven that the KL divergence is more effective than other metrics in training DNNs for predicting saliency.
To make the convergence speed faster, we pre-train the three branches. 
In particular, the visual and face branches are pre-trained on MVVA separately. For the visual branch, the RGB sub-branch is initialized with VGG parameters on ImageNet, while the Flow sub-branch is initialized with FlowNet parameters. The face branch is also initialized with VGG. Then, the audio branch is pre-trained jointly with the visual branch, since only audio cannot locate salient faces.

%\vspace{-5pt}
\section{Experiments and Results}
%\vspace{-3pt}

\subsection{Settings}
%\vspace{-4pt}
In our experiment, 300 videos in our MVVA are randomly divided into training (240 videos) and test (60 videos) sets. Specifically, for the visual branch, RGB frames are resized to 256x256. To train the convolutional LSTM, we temporally segment 240 training videos into 9,806 clips, all of which have $T = 12$ frames. 
For the audio branch, we use the 16-frame segmented log-mel spectrograms which are also resized to 256x256.
For the face branch, the resolution of $N$ input faces is 128x128.
The parameters of the proposed network are updated by using the Stochastic Gradient Descent (SGD) algorithm with Adam optimizer. The initial learning rate is set to be 1e-4.

To evaluate our method, we adopt four metrics: Area Under the receiver operating Characteristic curve (AUC), NSS, Correlation Coefficient (CC), and KL divergence \cite{borji2018saliency}. 
Note that the larger values for AUC, NSS or CC indicate more accurate saliency prediction. The opposite holds for the KL divergence. Please see \cite{bylinskii2018different} for more details on these metrics.
All experiments are conducted on a computer with Intel(R) Core(TM) i7-8700 CPU@3.20 GHz, 62.8 GB RAM and 2 Nvidia GeForce GTX 1080 Ti GPUs.

\begin{table}[t]
	%\tiny
	\centering
	%\vspace{-2.6em}
	\caption{Accuracy of saliency prediction by our method and 11 competing methods over different datasets.}
	%\vspace{-.1em}
	\resizebox{0.99 \textwidth}{!}{
		\begin{tabular}{c|cccccccccccccc}
			\toprule
			& & Ours & TASED & SAM\_res\hspace{-0.5em}& SAM\_vgg\hspace{-0.5em}& Liu\hspace{-0.5em}& ACLNet\hspace{-0.5em}& DeepVS \hspace{-0.5em}& SalGAN\hspace{-0.5em}& Coutrot & SALICON\hspace{-0.5em}& OBDL\hspace{-0.5em}& BMS & G-Eymol\hspace{-0.5em}\\
			\midrule
			\multirow{4}{*}{\tabincell{c}{ MVVA }} &AUC & \textbf{0.905} & 0.905 & 0.897 & 0.896 & 0.893 & 0.889 & 0.890 & 0.891 & 0.869 & 0.866 & 0.786 & 0.765 & 0.615\hspace{-0.5em}\\
			&NSS & \textbf{3.976} & 3.319 & 3.495 & 3.466 & 3.279 & 3.437 & 3.270 & 2.650 & 2.604 & 2.523 & 1.342 & 0.936 & 0.551\hspace{-0.5em}\\
			&CC & \textbf{0.722} & 0.653 & 0.634 & 0.634 & 0.625 & 0.639 & 0.615 & 0.539 & 0.509 & 0.477 & 0.273 & 0.193 & 0.125\hspace{-0.5em}\\
			&KL & \textbf{0.823} & 0.970 & 1.004 & 1.012 & 1.098 & 1.044 & 1.117 & 1.234 & 1.557 & 1.447 & 1.995 & 2.051 & 4.253\hspace{-0.5em}\\
			\midrule
			\midrule
			&AUC\hspace{-0.5em}& \textbf{0.922}\hspace{-0.5em}& 0.877 & 0.905\hspace{-0.5em}& 0.849\hspace{-0.5em}& 0.908\hspace{-0.5em}& 0.848 & 0.896 & 0.900 & 0.883 & 0.865 & 0.723 & 0.751 & 0.698\hspace{-0.5em}\\
			Coutrot II  &NSS\hspace{-0.5em}& \textbf{3.568}\hspace{-0.5em}& 2.731 & 3.446\hspace{-0.5em}& 3.306\hspace{-0.5em}& 2.833\hspace{-0.5em}& 3.127\hspace{-0.5em}& 3.058\hspace{-0.5em}& 2.286\hspace{-0.5em}& 3.033 & 2.408\hspace{-0.5em}& 0.730 & 0.739 & 0.884\hspace{-0.5em}\\
			\cite{coutrot2014saliency}&CC\hspace{-0.5em}& \textbf{0.639}\hspace{-0.5em}& 0.545 & 0.607\hspace{-0.5em}& 0.593\hspace{-0.5em}& 0.585\hspace{-0.5em}& 0.521\hspace{-0.5em}& 0.556\hspace{-0.5em}& 0.553\hspace{-0.5em}& 0.606 & 0.433\hspace{-0.5em}& 0.181 & 0.153 & 0.162\hspace{-0.5em}\\
			&KL\hspace{-0.5em}& \textbf{0.915}\hspace{-0.5em}& 1.271 & 1.031\hspace{-0.5em}& 1.093\hspace{-0.5em}& 1.035\hspace{-0.5em}& 1.357\hspace{-0.5em}& 1.209\hspace{-0.5em}& 1.717\hspace{-0.5em}& 1.428 & 1.514\hspace{-0.5em}& 2.228 & 2.073 & 2.932\hspace{-0.5em}\\
			\bottomrule
	\end{tabular}}%
	\label{tab:acc_compare}%
	%\vspace{-3.6em}
\end{table}%

\subsection{Performance Comparison}
%\vspace{-3pt}
We compare the performance of our multi-modal method with 11 state-of-the-art saliency prediction methods, including TASED \cite{min2019tased}, SAM \cite{cornia2018predicting}, Liu \cite{liu2017predicting}, ACLNet \cite{wang2018revisiting}, DeepVS \cite{jiang_OMCNN}, SalGAN \cite{pan2017salgan}, SALICON \cite{huang2015salicon}, Coutrot \cite{coutrot2015efficient}, OBDL \cite{hossein2015many}, BMS \cite{zhang2016exploiting} and G-Eymol \cite{zanca2019gravitational}. Among them, SalGAN, SALICON, SAM and BMS are state-of-the-art saliency prediction methods over images, and others are for videos. SAM has two versions, including SAM\_res with ResNet backbone and SAM\_vgg with VGGNet backbone. Note that Coutrot and Liu focus on multiple-face videos.
In Coutrot, static saliency map, dynamic saliency map, speaker map and center bias map are weighted with estimated weights, and merged into the final saliency map.
To eliminate the influence of the feature extraction algorithm (e.g., face/speaking detection), we re-implement Coutrot \etal method with manual annotated features and treat the performance as the upper bound of Coutrot \etal
Liu is the latest DL based method for multiple-face videos, but it ignores the audio information.
Besides, face is also considered in G-Eymol as a semantic-based feature.
To effectively assess the power of our method, we test it on different databases as follows.

\begin{figure}[t]
	\begin{center}
		%\vspace{-.4em}
		%\hspace{-1.4em}
		\includegraphics[width=1\linewidth]{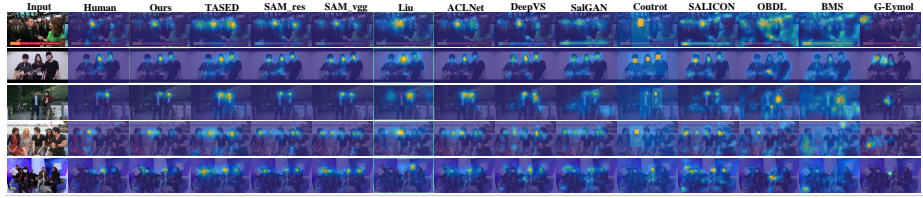}
	\end{center}
	%\vspace{-2.1em}
	\caption{Saliency maps of 5 videos randomly selected from the test set of our eye-tracking database.}
	\label{fig:vis_result1}
	%\vspace{-.8em}
\end{figure}

\begin{figure*}[t]
	\begin{center}
		%\vspace{-.6em}
		\hspace{-2em}
		\includegraphics[width=1\linewidth]{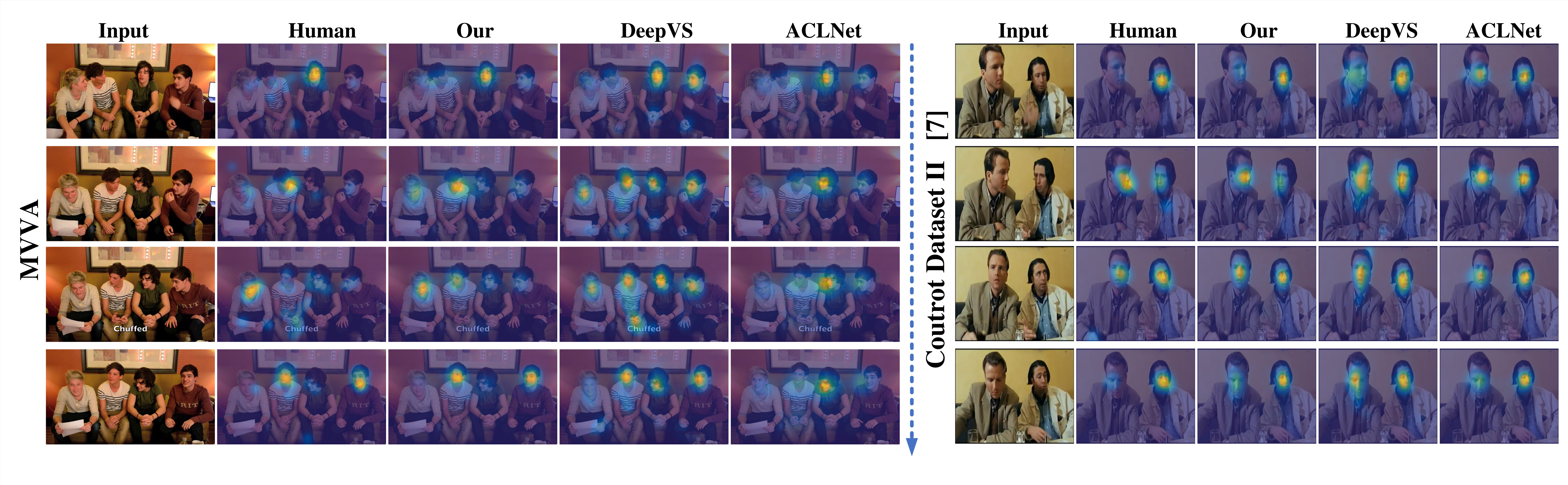}
	\end{center}
	%\vspace{-2.3em}
	\caption{Saliency maps for different frames of two video sequences, selected from our MVVA and Coutrot II \cite{coutrot2014saliency}.}
	\label{fig:vis_result2}
	%\vspace{-2.2em}
\end{figure*}

\noindent\textbf{Evaluation on our dataset.}
Tab. \ref{tab:acc_compare} presents AUC, NSS, CC and KL divergence for the proposed method versus 11 competing methods. Scores are averaged over 60 test videos in our eye-tracking database. As shown in this table, the proposed method performs significantly better than all other methods over all 4 metrics. Specifically, compared with the best competing result, our method achieves over 0.481, 0.069 and 0.147 improvements in NSS, CC and KL, respectively. The main reasons for this result are: 1) Most of state-of-the-art methods do not consider audio information, while our method does utilize audio cue for saliency prediction, 2) The face temporal subnet of our method learns detailed face features to predict salient faces, and 3) Our fusion module effectively integrates the multi-modal information.

Next, we compare models qualitatively. Fig. \ref{fig:vis_result1} demonstrates saliency maps over 5 randomly selected videos in the test set, predicted by the proposed method and 11 other methods.
As is shown, our method is capable of locating the salient faces. Its prediction is much closer to the GT.
Besides, the proposed method shows excellent performance on predicting attention transition, as depicted in Fig. \ref{fig:vis_result2}.
In contrast, most of the other methods fail to accurately predict the regions that attract human attention, perhaps because these methods do not consider extra information such as sound and face.

\noindent\textbf{Evaluation on generalization ability.}
To evaluate the generalization capability of the proposed method, we further evaluate our method and 11 other methods on the Coutrot II database \cite{coutrot2014saliency}. Tab. \ref{tab:acc_compare} compares the average AUC, NSS, CC and KL scores. As shown in this table, the proposed method again outperforms all the competing methods. In particular, there are at least 0.032 and 0.116 improvements in CC and KL, respectively. Such improvements are comparable to those in our MVVA.
Qualitative results, shown in Fig. \ref{fig:vis_result2}, shows the proposed method predicts attention transition accurately, while other methods miss salient faces.
These results demonstrate the generalization capability of our method in video saliency prediction.

\begin{table}[t]
	%\scriptsize
	\centering
	%\vspace{-2.3em}
	\caption{Performance of different modules in our model.}
	%\vspace{-0.8em}
	\resizebox{0.8 \textwidth}{!}{
		\begin{tabular}{c|ccccc}
			\toprule
			  &\hspace{.8em} Models \hspace{.8em}&\hspace{.8em} CC \hspace{.8em}&\hspace{.8em} KL \hspace{.8em}&\hspace{.8em} NSS \hspace{.8em}&\hspace{.8em} AUC \\
			\midrule
			%\multirow{2}{*}{\tabincell{c}{ Bounds }} &\hspace{.8em} Avg. baseline (lower bound) &\hspace{.8em} 0.364 \hspace{.8em}&\hspace{.8em} 1.575 \hspace{.8em}&\hspace{.8em} 1.614 \hspace{.8em}&\hspace{.8em} 0.848 \\
			%&Human (upper bound) \hspace{.8em}&\hspace{.8em} 0.747 \hspace{.8em}&\hspace{.8em} 1.278 \hspace{.8em}&\hspace{.8em} 4.573 \hspace{.8em}&\hspace{.8em} 0.875 \\
			%\midrule
			\multirow{8}{*}{\tabincell{c}{ Different modules }} 
			& Visual (RGB only) &\hspace{.8em} 0.527 \hspace{.8em}&\hspace{.8em} 1.324 \hspace{.8em}&\hspace{.8em} 2.728 \hspace{.8em}&\hspace{.8em} 0.860\\
			& Visual (Flow only) &\hspace{.8em} 0.510 \hspace{.8em}&\hspace{.8em} 1.354 \hspace{.8em}&\hspace{.8em} 2.631 \hspace{.8em}&\hspace{.8em} 0.869\\
			& Visual (RGB+flow) &\hspace{.8em} 0.632 \hspace{.8em}&\hspace{.8em} 1.043 \hspace{.8em}&\hspace{.8em} 3.358 \hspace{.8em}&\hspace{.8em} 0.893 \\
			& Visual (RGB+flow+LSTM) &\hspace{.8em} 0.671 \hspace{.8em}&\hspace{.8em} 0.971 \hspace{.8em}&\hspace{.8em} 3.548 \hspace{.8em}&\hspace{.8em} 0.896 \\
			& Visual+audio  &\hspace{.8em} 0.712 \hspace{.8em}&\hspace{.8em} 0.843 \hspace{.8em}&\hspace{.8em} 3.838 \hspace{.8em}&\hspace{.8em} 0.907 \\
			& Face only  &\hspace{.8em}  0.569 \hspace{.8em}&\hspace{.8em} 1.292 \hspace{.8em}&\hspace{.8em} 2.766 \hspace{.8em}&\hspace{.8em} 0.872\\
			& Face+audio &\hspace{.8em} 0.609 \hspace{.8em}&\hspace{.8em} 1.116 \hspace{.8em}&\hspace{.8em} 3.211 \hspace{.8em}&\hspace{.8em} 0.878\\
			& Visual+audio+face &\hspace{.8em} \textbf{0.722} \hspace{.8em}&\hspace{.8em} \textbf{0.823} \hspace{.8em}&\hspace{.8em} \textbf{3.976} \hspace{.8em}&\hspace{.8em} \textbf{0.905} \\
			%\midrule
			%\multirow{3}{*}{\tabincell{c}{ Different\\ \hspace{.8em}combination tectniques\hspace{.8em} }} & Concatenation &\hspace{.8em} 0.722 \hspace{.8em}&\hspace{.8em} 0.823 \hspace{.8em}&\hspace{.8em} 3.976 \hspace{.8em}&\hspace{.8em} 0.905 \\
			%&Element-wise sum &\hspace{.8em} 0.704 \hspace{.8em}&\hspace{.8em} 0.875 \hspace{.8em}&\hspace{.8em} 3.854 \hspace{.8em}&\hspace{.8em} 0.901 \\
			%&Element-wise product  &\hspace{.8em} 0.699 \hspace{.8em}&\hspace{.8em} 0.893 \hspace{.8em}&\hspace{.8em} 3.781 \hspace{.8em}&\hspace{.8em} 0.906 \\
			\bottomrule
	\end{tabular}}%
	\label{tab:ablation}%
	%\vspace{-1.8em}
\end{table}%

%\vspace{-6pt}
\subsection{Ablation Analysis}
%\vspace{-3pt}
Here, we thoroughly analyze the effectiveness of each module in our method.

\noindent\textbf{Visual branch}.
Visual branch uses basic visual information, i.e., texture, motion and temporal cues, to predict saliency. We evaluate the visual branch of the proposed network and report the results in Tab. \ref{tab:ablation}. It shows that visual branch reaches to CC of 0.632 and KL of 1.043, which is better than many methods and comparable with the best competing method TASED. When adding convolutional LSTM to fuse the temporal cues, the performance reaches to 0.671 in CC and 0.971 in KL. Hence, the entire visual branch and its components are all useful to saliency prediction.
Moreover, as shown in Tab. \ref{tab:ablation}, combination of face and audio results in lower performance than combining all cues (i.e., the whole network, visual+audio+face) by a large margin. It further manifests the effectiveness of visual branch. We add visual branch, because there are still some other regions drawing attention, besides faces.

\noindent\textbf{Audio branch}.
Besides visual branch, we add audio branch to the framework.
With the help of the audio branch, the visual-audio model achieves 0.712 in CC and 0.843 in KL, much better than the visual branch. In addition, the combination of face and audio branches improves the performance of the single face branch, by 0.040 in CC and 0.176 in KL.
Thus, these results manifest the contribution of audio information and the effectiveness of the proposed audio branch.

%\begin{comment}
\begin{figure}[t]
	\begin{minipage}{0.7\linewidth}
	\centering
		%\vspace{-.1em}
		\includegraphics[width=.8\linewidth]{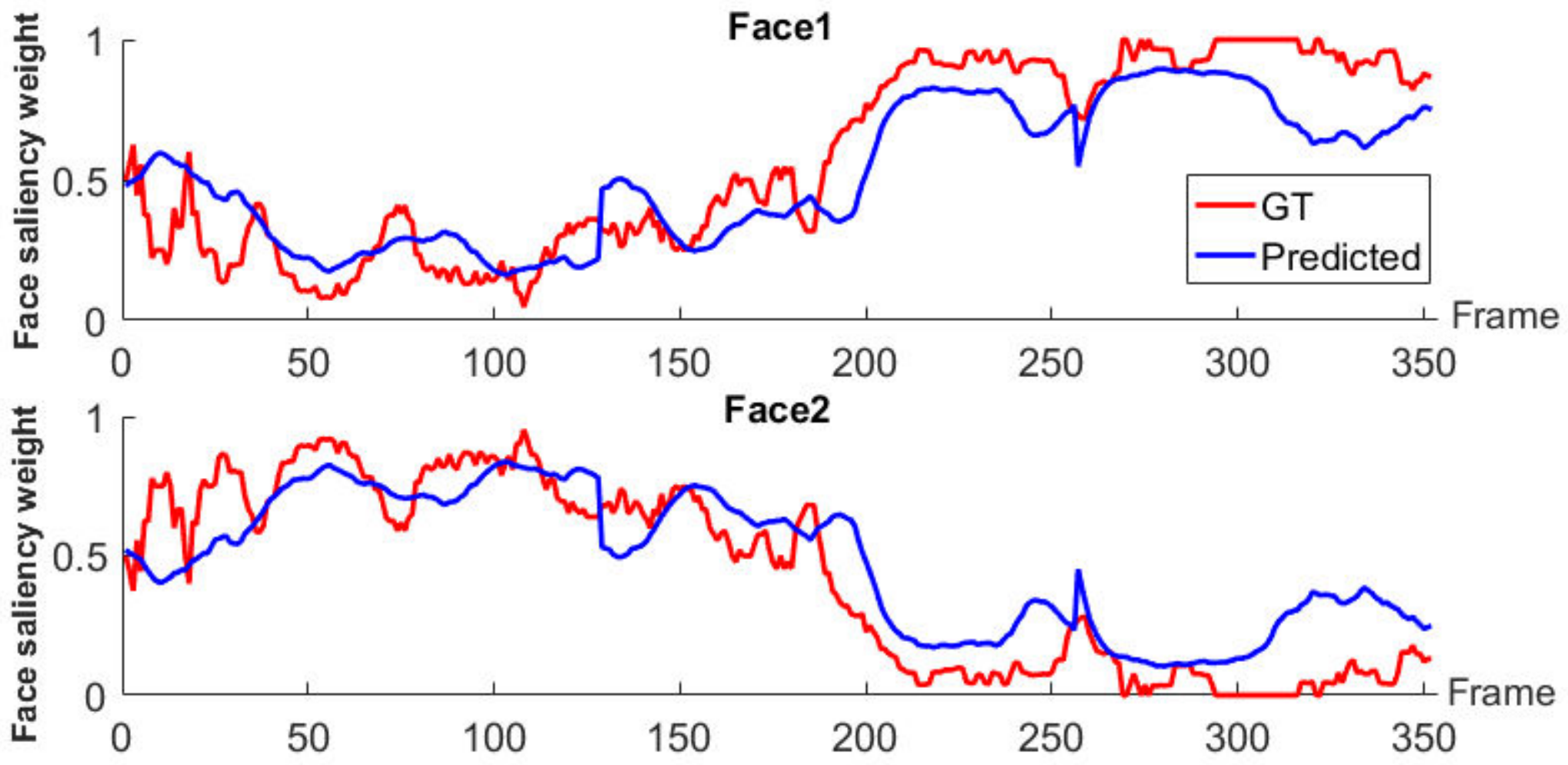}
	\end{minipage}
     \hfill
	%\vspace{-1.8em}
	\hspace{-.6em}
	\begin{minipage}{0.3\linewidth}
	\centering
	\caption{Face saliency weights across frames for a randomly selected video.}
	\label{fig:face_weight_curve}
    \end{minipage}
	%\vspace{-2.2em}
	%\label{fig:face_weight_curve}
\end{figure}
%\end{comment}

\noindent\textbf{Face branch}.
Finally, the face branch is added to complete the whole network. From Tab. \ref{tab:ablation}, CC of 0.722 and KL of 0.823 are reached, after combining face branch with visual-audio model. It is worth mentioning that the single face branch can only achieve a fair performance, which is inferior to other combinations. Hence, single face branch cannot reach the best accuracy, even most attention is attracted by faces.
In addition, since the face branch aims at predicting saliency weight of faces across the video frames, we plot the face saliency weights of the proposed face branch and GT in Fig. \ref{fig:face_weight_curve}. In this figure, the curve of the face branch fits close to the curve of GT. 
It can be concluded that the face branch accurately predicts the salient face and further enhances the performance of the proposed model.

In summary, the ablation analysis manifests the necessity of different cues for saliency prediction, and verifies the effectiveness of each part in our model.  
More details can be found in the supplementary document.

%\vspace{-6pt}
\section{Conclusion}
%\vspace{-5pt}
In this paper, we explored how audio influences human attention in multiple-face videos.
Various findings have been verified by the statistical analysis on our new eye-tracking database.
To predict multiple-face video saliency, we presented a novel multi-modal network consisting of visual, audio and face branches. The three branches encode visual frames, audio spectrograms and faces into feature maps, respectively. A fusion module was designed to integrate the three modalities, and to generate the final saliency map. Finally, experimental results shown that our method outperforms 11 state-of-the-art methods over several datasets.

\noindent\textbf{Acknowledgement.} This work is supported by Beijing Natural Science Foundation (Grant No. L172051, JQ18018), the Natural Science Foundation of China (Grant No. 61902401, 61972071, 61751212, 61721004, 61876013, 61922009, 61573037 and U1803119), the NSFC-general technology collaborative Fund for basic research (Grant No. U1636218, U1936204),  CAS Key Research Program of Frontier Sciences (Grant No. QYZDJ-SSW-JSC040), CAS External cooperation key project, and NSF of Guangdong (No. 2018B030311046). Bing Li is also supported by CAS Youth Innovation Promotion Association.

%\clearpage

\bibliographystyle{splncs04}
\bibliography{egbib}
\end{document}